\documentclass[a4paper,conference]{IEEEtran}
\usepackage{threeparttable}
\usepackage{ifthen}
\usepackage{color}
\usepackage{keyval}
\usepackage{multirow}

\usepackage{ifpdf}

\usepackage{cite}

   \usepackage[pdftex]{graphicx}
\usepackage{amsmath}
\usepackage{algorithmic}

\usepackage{array}

  \usepackage[caption=false,font=normalsize,labelfont=sf,textfont=sf]{subfig}
\usepackage{fixltx2e}

\usepackage{stfloats}
\begin{document}
%
\title{ Co-occurrence Background Model with Superpixels for Robust Background Initialization 
}



%
\author{\IEEEauthorblockN{Wenjun Zhou\IEEEauthorrefmark{1},
Yuheng Deng\IEEEauthorrefmark{1},
Bo Peng\IEEEauthorrefmark{1},
Dong Liang\IEEEauthorrefmark{2} and
Shun'ichi Kaneko\IEEEauthorrefmark{3}}
\IEEEauthorblockA{\IEEEauthorrefmark{1}School of Computer Science, Southwest Petroleum University, Chengdu, China 610500\\ Email: zhouwenjun@swpu.edu.cn, dengyuhengSWPU@outlook.com}
\IEEEauthorblockA{\IEEEauthorrefmark{2}Nanjing University of Aeronautics and Astronautics, China}
\IEEEauthorblockA{\IEEEauthorrefmark{3}Hokkaido University, Japan}}


\maketitle

\begin{abstract}
Background initialization is an important step in many high-level applications of video processing, ranging from video surveillance to video inpainting. However, this process is often affected by practical challenges such as illumination changes, background motion, camera jitter and intermittent movement, etc. In this paper, we develop a co-occurrence background model with superpixel segmentation for robust background initialization. We first introduce a novel co-occurrence background modeling method called as Co-occurrence Pixel-Block Pairs (CPB) to generate a reliable initial background model, and the superpixel segmentation is utilized to further acquire the spatial texture information of foreground and background. Then, the initial background can be determined by combining the foreground extraction results with the superpixel segmentation information. Experimental results obtained from the dataset of the challenging benchmark (SBMnet) validate it's performance under various challenges.
\end{abstract}


%
\IEEEpeerreviewmaketitle

\section{Introduction}\label{sec:intro}
\begin{figure}[!htb]
	\centering
	\includegraphics[width=0.4\textwidth]{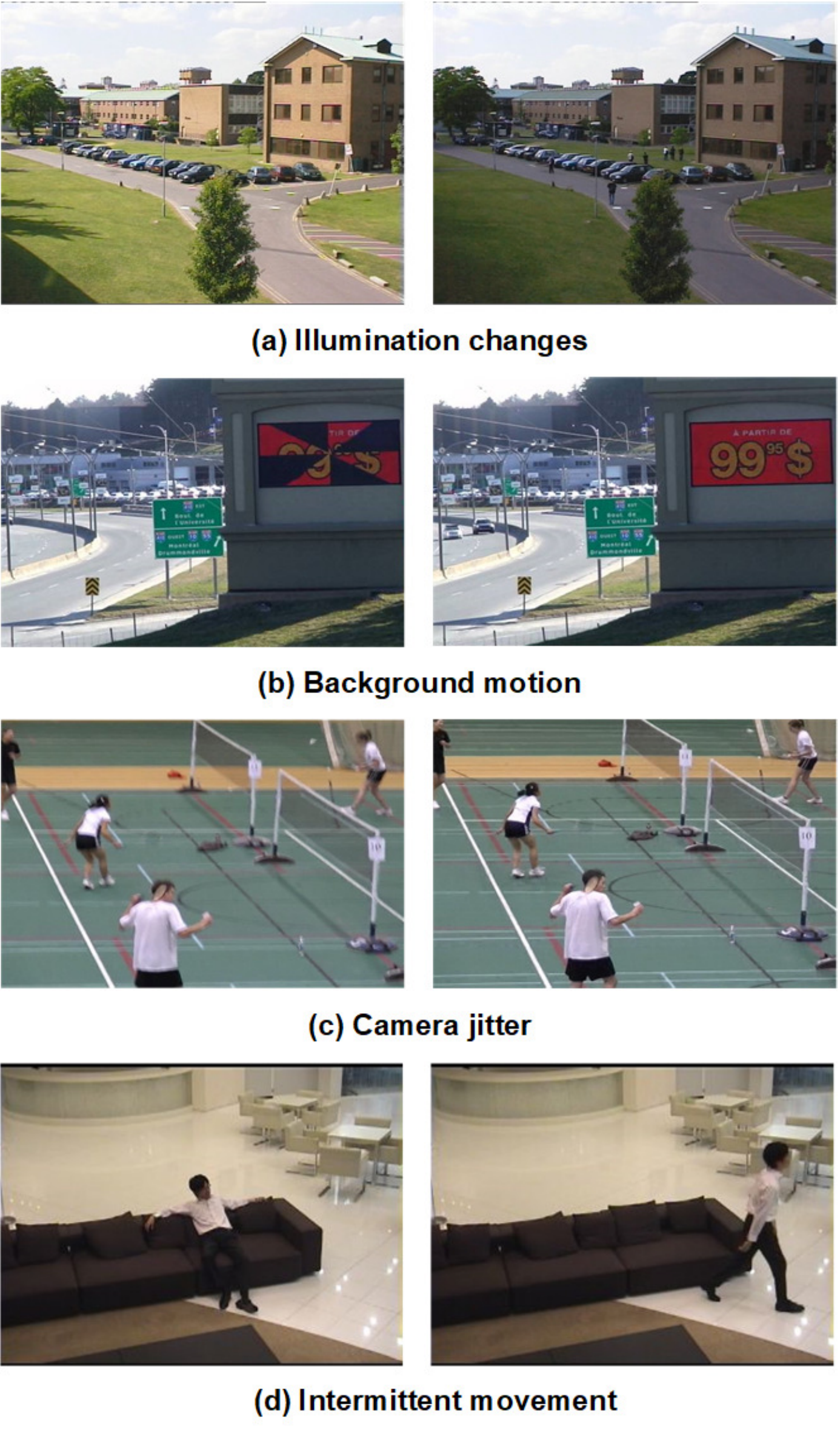}
	\caption{Typical examples of these challenges: (a) Illumination changes, (b) Background motion, (c) Camera jitter, (d) Intermittent motion.}
	\label{fig:fig1}
\end{figure}

As a widely used approach in various computer vision and video processing applications\cite{bouwmans2014traditional, bouwmans2017scene}, scene background initialization plays an active role in object detection\cite{zhang2017a}, video segmentation\cite{chiu2010a}, video coding\cite{paul2012efficient, li2016background} and video inpainting\cite{colombari2005exemplar, chen2010background}, etc. Scene background initialization describes the scene without any foreground objects and generates a clear background to facilitate more efficient follow-up processing in computer vision or video processing applications. Bouwmans et al. overviewed and summarized many traditional and recent approaches that have been proposed and developed for scene background initialization\cite{bouwmans2017scene}, and previous works\cite{maddalena2015towards,jodoin2017extensive} have already analyzed challenges of background initialization. However, background initialization is still faced with some severe practical challenges\cite{xu2019robust} which include:

\begin{itemize}
	\item \textbf{Illumination changes:} for example, light intensity typically varies during day. 
	
	\item \textbf{Background motion:} some movements in a scene should be determined as background e.g. swaying tree, waving water, or ever-changing advertising boards. 
	
	\item \textbf{Camera jitter:} in video surveillance, camera jitter is one severe issue that needs to be solved for background initialization.
	
	
	\item \textbf{Intermittent movement:} the scene with abandoned objects stopping for a short while and then moving away. Under this condition, to differentiate between foreground and abandoned objects is difficult.

\end{itemize}
Fig.~\ref{fig:fig1} shows the typical examples of these challenges.

To handle above challenges, we propose a robust background initialization approach based on the co-occurrence background model (Co-occurrence Pixel-Block Pairs: CPB) with superpixels. CPB has already been described in our previous work\cite{zhou2017visual,zhou2019foreground}. As an intuitive and robust background model, CPB was originally designed for foreground detection under dramatical background changes, such as illumination changes and background motion. Here, CPB is utilized as the background model for scene background initialization. Then, in order to further obtain the spatial texture information of foreground and background for efficient background generation, the superpixel algorithm called simple linear iterative clustering (SLIC)  \cite{achanta2012slic} is introduced to classify the spatial correlations and temporal difference motion  between foreground and background for motion detection. The main contributions of this work are as follows: 

\begin{enumerate}
	\item[1.] The proposed approach enables to effectively acquire the spatial-temporal information of foreground and background and sensitively distinguish the difference between them, so it is highly efficient for motion detection in a scene under complex challenges, especially strong background changes (e.g. \textit{illumination changes} and \textit{background motion}) or \textit{intermittent motion}. 
	\item[2.] The proposed approach provides a low-complexity and efficient strategy for robust background initialization. Especially when compared with neural network (\textit{NN}) based approaches\cite{xu2014motion, halfaoui2016cnn}, it has low cost because it is capable of training without any teacher signals.
	
\end{enumerate} 

The rest of this paper is organized as follows. The proposed approach is described in Section~\ref{sec:meth}. 
Section~\ref{sec:experimental} analyzes the experimental results from  the dataset of the SBMnet\cite{SBMnet2016}. Conclusions are discussed in Section~\ref{sec:conclusion}.

\section{Methodology}\label{sec:meth}
\begin{figure*}[!htb]
	\centering
	\includegraphics[width=0.9\textwidth]{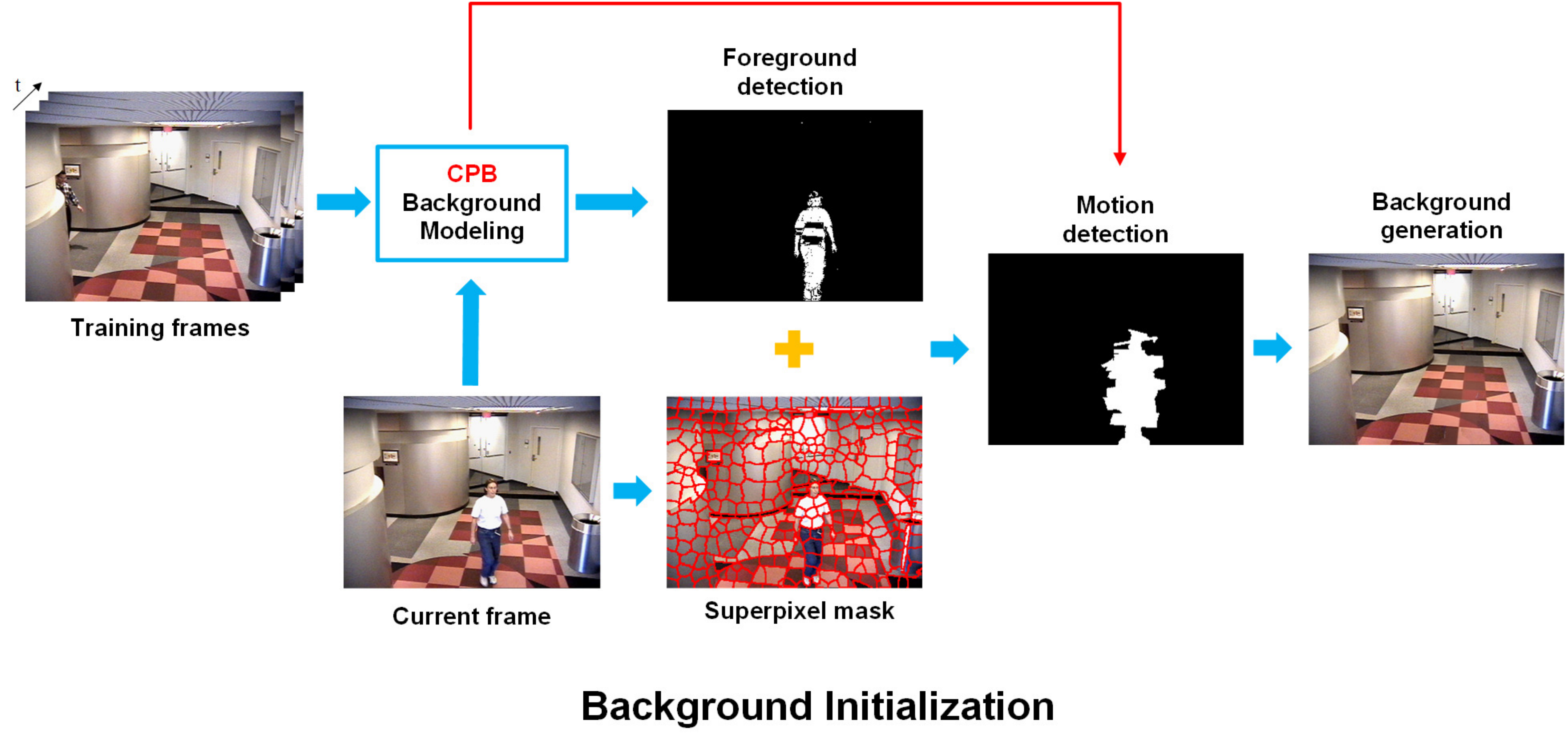}
	\caption{Overview of background initialization by the proposed approach.}
	\label{fig:fig3}
\end{figure*}

In this section, the proposed approach is described in details. The steps of it includes: (1) CPB background modeling; (2) Motion detection; (3) Background generation as shown in Fig.~\ref{fig:fig3}. 

\subsection{Co-occurrence Background Model}
\label{ssec:co-occurrence}
\begin{figure*}[!htb]
	\centering
	\includegraphics[width=0.9\textwidth]{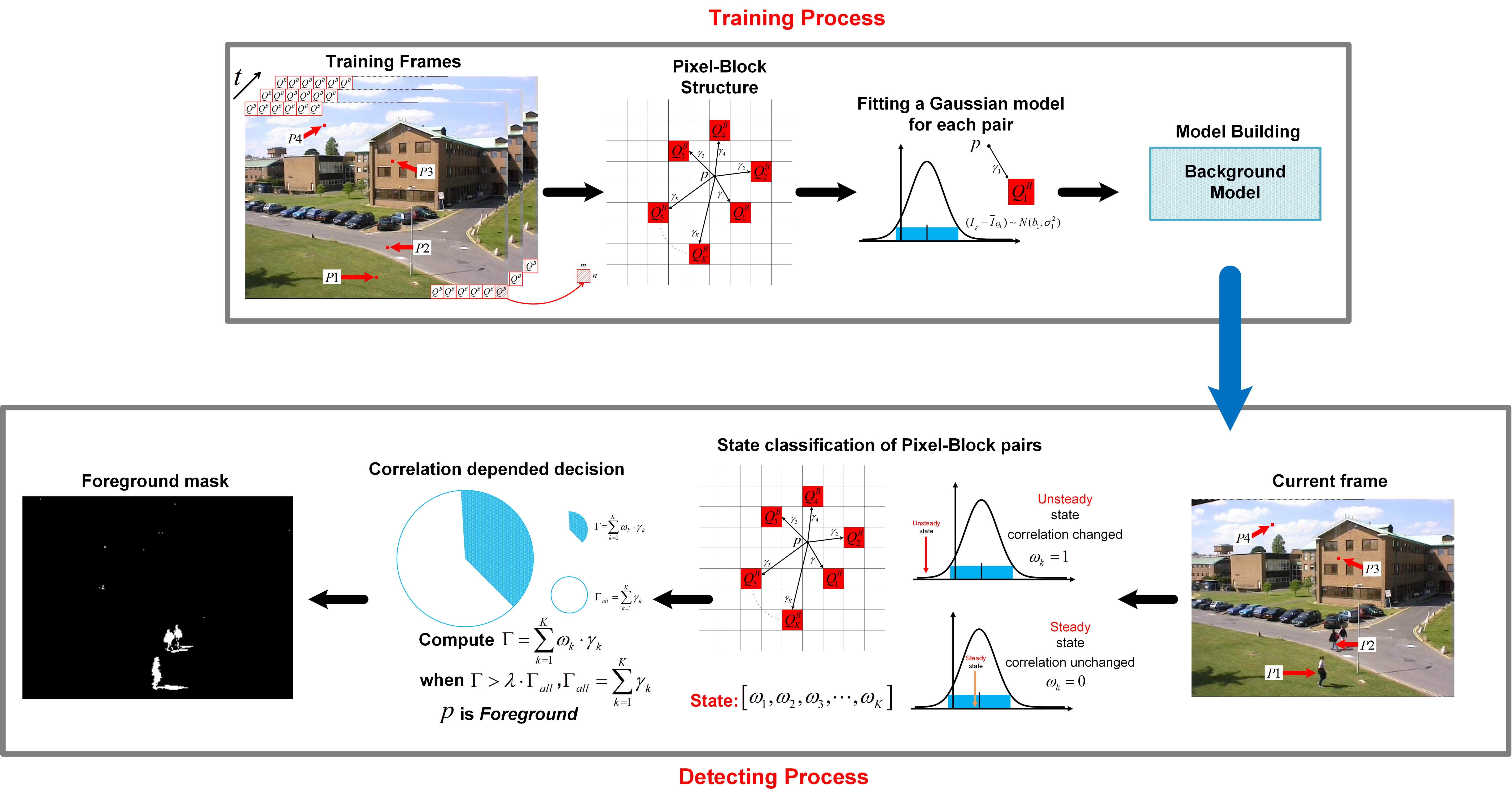}
	\caption{Working diagram of CPB background model using PETS2001 dataset as a demonstration.}
	\label{fig:fig2}
\end{figure*}

The working diagram of CPB background modeling is illustrated in Fig.~\ref{fig:fig2} including: the training process and the detecting process. In this work, the target pixel $p$ is compared with the $Q^B$ as block, and we define $\{Q_k^B\}_{k=1,2,...,K}=\{Q_1^B,Q_2^B,...,Q_K^B\}$ to denote a supporting block set for the target pixel $p$.
Each frame is divided into blocks $Q_K^B$ of size $m \times n$ pixels:

\begin{equation}\label{Q^B}
Q^B=
\begin{Bmatrix}
Q_{11} & Q_{12} & \dots & Q_{1n}  \\
Q_{21} & Q_{22} & \dots & Q_{2n}  \\
\vdots & \vdots & \vdots & \vdots \\
Q_{m1} & Q_{m2} & \dots & Q_{mn}
\end{Bmatrix}.
\end{equation}

Background changes in scene can affect the current intensity of target pixel $p$ in foreground detection. Hence, it is quite natural that block $Q^B$, being strongly correlated with target pixel $p$, can be used to determine the state of the latter. Block $Q^B$ can be introduced as a reference to estimate the current intensity of target pixel $p$, that is, there exists a correlation between pixel $p$ and block $Q^B$: $I_p= \bar  I_Q +\Delta_k$ ($ \bar I_Q$ is the average intensity of block $Q^B$ in the current detecting frame). In order to reduce the risk of individual error and perform robust background model, to select the sufficient
number of  block $Q^B$ with high correlation as supporting blocks is necessary, defined as follows:

\begin{equation}\label{r}
\{Q_k^B\}_{k=1,2,...,K}=\{Q^B|\gamma(p,Q^B) \text{is the $K$  highest}\},
\end{equation}
where

\begin{equation}\label{R}
{\gamma}(p,Q_k^B)=\dfrac{C_{p ,\bar Q_k}}{\sigma_{p}\cdot\sigma_{\bar Q_k}},
\end{equation}
where $\gamma$ is the Pearson’s product-moment correlation coefficient. Then, the Gaussian model is used to construct the co-occurrence model for each pixel-block pair:

\begin{equation}\label{Dela}
\Delta_{k}\sim N(b_k,\sigma_k^2) \quad \Delta_{k}=I_p-\bar I_{Q_k},
\end{equation}
where $ I_{p}$ is the intensity of the pixel $p$ at $t$ frame and $\bar I_{Q_k}$ is the average intensity of blocks $Q_k^B$ at $t$ frame. The background model is built as a list consisting of $[I^{P},u_k,v_k,b_k,\sigma_k]$, where $I^{P}$ is the average intensity of target pixel $p$ in $T$ sequence frames computed by training ($T$ is the number of training frames) and $(u_k,v_k)$ are the coordinates of supporting blocks.

At the detecting process, we use the correlation dependent decision for identifying the state of target pixel $p$ as shown in Fig.\ref{fig:fig2} and more details are described in\cite{zhou2019foreground}.

\begin{figure*}[!htb]
	\centering
	\includegraphics[width=0.88\textwidth]{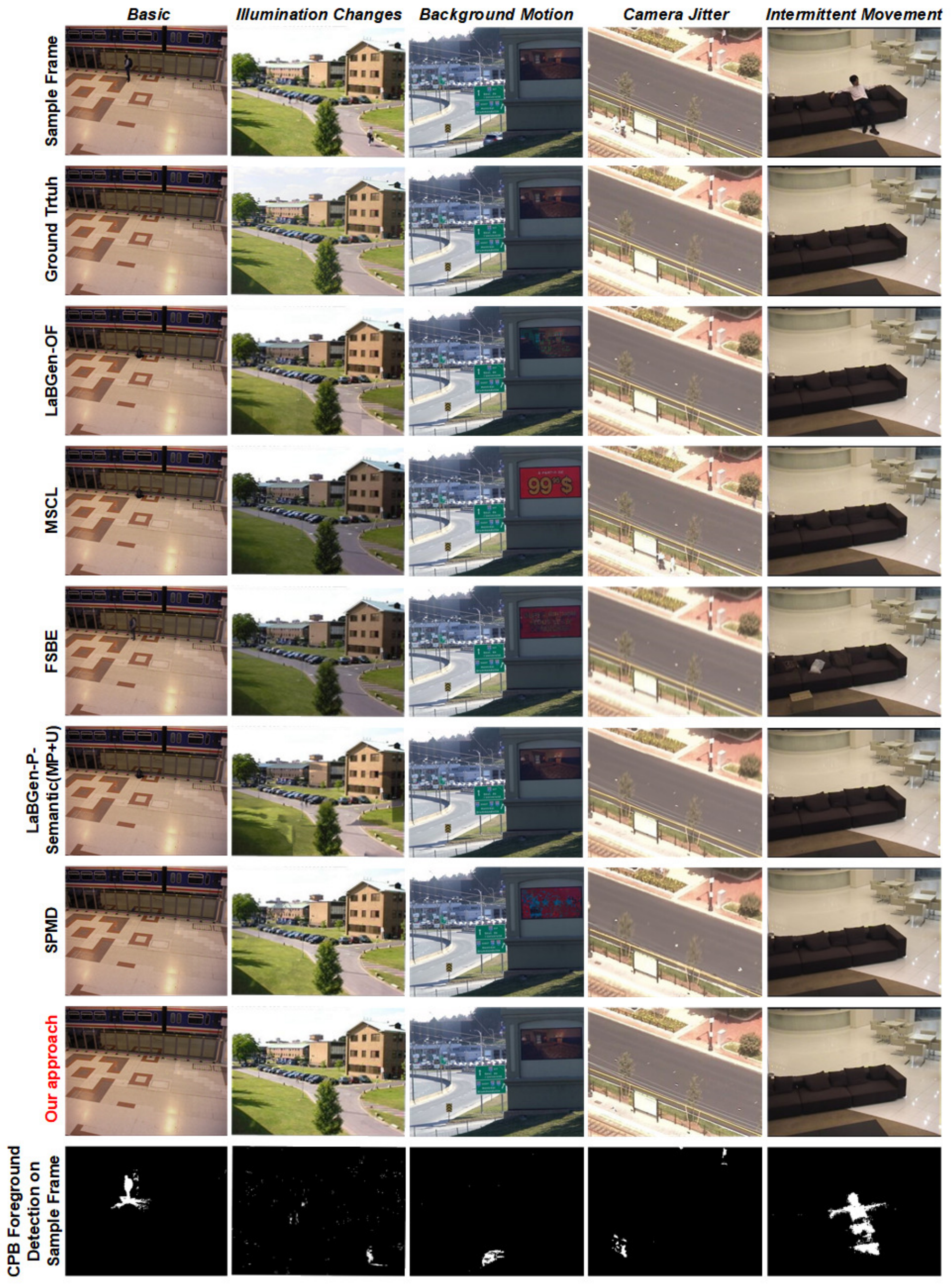}
	\caption{Representative results in different challenging sequences.}
	\label{fig:fig4}
\end{figure*}

\begin{table*}[!htb]
	
	\renewcommand{\arraystretch}{}
	\caption{Results on different challenges from the \textit{SBMnet} dataset}
	\label{Tab:Tab3}
	\centering
	\scalebox{1.0}[1.0]{
		\begin{threeparttable}
			\begin{tabular}{cccccccc}
				
				\hline 
				Challenge                                                                         & Method                                                             & AGE                        & pEPs                       & pCEPs                      & PSNR                   & MS-SSIM                     & CQM                         \\ \hline 
				& LaBGen-OF                                                          & 1.8388                     & 0.0026                     & 0.0017                     & 0.9899                     & 34.6563                     & 35.4184                     \\
				& MSCL                                                               & 2.3728                     & 0.0027                     & 0.0016                     & 0.9866                     & 34.081                      & 34.7595                     \\
				& FSBE                                                               & 3.0236                     & 0.0055                     & 0.0035                     & 0.9821                     & 33.6317                     & 34.2344                     \\
				& \begin{tabular}[c]{@{}c@{}}LaBGen-P-Semantic\\ (MP+U)\end{tabular} & 1.9743                     & 0.0024                     & 0.0015                     & 0.9899                     & 34.8111                     & 35.5647                     \\
				& SPMD                                                               & 2.1919                     & 0.0004                     & 0.0000                     & 0.9935                     & 38.6807                     & 38.9381                     \\
				\multirow{-6}{*}{Basic}                                                           & \textbf{Our approach}                                & \textcolor[rgb]{1,0,0}{1.4275}                     & \textcolor[rgb]{1,0,0}{0.0002}                     & \textcolor[rgb]{1,0,0}{0.0000}                     &\textcolor[rgb]{1,0,0} {0.9983}                     & \textcolor[rgb]{1,0,0}{42.3151}                     & \textcolor[rgb]{1,0,0}{42.2216}                     \\\hline
				& LaBGen-OF                                                          & 19.6355                    & 0.4062                     & 0.2597                     & 0.9346                     & 19.4204                     & 20.9417                     \\
				& MSCL                                                               & \textcolor[rgb]{1,0,0}{2.8098}                     & \textcolor[rgb]{1,0,0}{0.0043}                     & \textcolor[rgb]{1,0,0}{0.0000}                     & \textcolor[rgb]{1,0,0}{0.9913}                     & \textcolor[rgb]{1,0,0}{34.9208}                     & \textcolor[rgb]{1,0,0}{35.5259}                     \\
				& FSBE                                                               & 6.6733                     & 0.0177                     & 0.0002                     & 0.9817                     & 29.2464                     & 30.2773                     \\
				& \begin{tabular}[c]{@{}c@{}}LaBGen-P-Semantic\\ (MP+U)\end{tabular} & 17.6197                    & 0.2733                     & 0.1829                     & 0.8641                     & 18.4939                     & 20.083                      \\
				& SPMD                                                               & 6.0889                     & 0.0540                     & 0.0129                     & 0.9755                     & 26.9955                     & 28.1438                     \\
				\multirow{-6}{*}{\begin{tabular}[c]{@{}c@{}}Illumination\\ Changes\end{tabular}}  & \textbf{Our approach}                                & 15.2618                    & 0.1657                     & 0.0130                     & 0.9451                     & 21.3651                     & 22.3365                     \\\hline
				& {LaBGen-OF}                                   & 1.7604                     & 0.0022                     & 0.0005                     & 0.9893                     & 38.6184                     & 39.0805                     \\
				& MSCL                                                               & 2.1299                     & 0.0016                     & 0.0005                     & 0.9962                     & 36.6006                     & 36.8315                     \\
				& FSBE                                                               & 1.8453                     & 0.0029                     & 0.0003                     & 0.9814                     & 37.9984                     & 37.9817                     \\
				& \begin{tabular}[c]{@{}c@{}}LaBGen-P-Semantic\\ (MP+U)\end{tabular} & \textcolor[rgb]{1,0,0}{1.5156}                     & \textcolor[rgb]{1,0,0}{0.0000}                    & \textcolor[rgb]{1,0,0}{0.0000}                     & \textcolor[rgb]{1,0,0}{0.9970}                     & \textcolor[rgb]{1,0,0}{41.4472}                     & \textcolor[rgb]{1,0,0}{41.4719}                     \\
				& SPMD                                                               & 2.2313                     & 0.0035                     & 0.0002                     & 0.9823                     & 36.8531                     & 36.1390                      \\
				\multirow{-6}{*}{\begin{tabular}[c]{@{}c@{}}Background \\ Motion\end{tabular}}    & \textbf{Our approach}                                & 1.7742                     & \textcolor[rgb]{1,0,0}{0.0000}                     & \textcolor[rgb]{1,0,0}{0.0000}                     & 0.9965                     & 39.7339                     & 39.9130                     \\\hline
				& LaBGen-OF                                                          & 11.9868                    & 0.1590                     & 0.0267                     & 0.8719                     & 20.2275                     & 21.7778                     \\
				& MSCL                                                               & 5.8660                     & 0.0471                     & 0.0067                     & 0.9699                     & 26.0077                     & 27.1642                     \\
				& FSBE                                                               & 10.1060                    & 0.1413                     & 0.0283                     & 0.9003                     & 22.5280                     & 23.8107                     \\
				& \begin{tabular}[c]{@{}c@{}}LaBGen-P-Semantic\\ (MP+U)\end{tabular} & 11.1637                    & 0.1466                     & 0.0281                     & 0.8619                     & 20.4535                     & 21.8627                     \\
				& SPMD                                                               & \textcolor[rgb]{1,0,0}{1.3573}                     & \textcolor[rgb]{1,0,0}{0.0001}                     &\textcolor[rgb]{1,0,0}{0.0000}                     & \textcolor[rgb]{1,0,0}{0.9979}                     &\textcolor[rgb]{1,0,0}{42.1226}                     & \textcolor[rgb]{1,0,0}{42.1988}                     \\
				\multirow{-6}{*}{\begin{tabular}[c]{@{}c@{}}Camera\\ Jitter\end{tabular}}         & \textbf{Our approach}                                                       & 9.4038                     & 0.1205                     & 0.0133                     & 0.9235                     & 22.6436                     & 24.0308                     \\\hline
				& LaBGen-OF                                                          & 2.3248                     & 0.0043                     & 0.0021                     & 0.9948                     & 36.5121                     & 36.8640                     \\
				& MSCL                                                               & 1.8481                     & 0.0026                     & 0.0011                     & 0.9943                     & 37.9796                     & 38.1597                     \\
				& FSBE                                                               & 3.8068                     & 0.0263                     & 0.0173                     & 0.9432                     & 27.9022                     & 28.9156                     \\
				& \begin{tabular}[c]{@{}c@{}}LaBGen-P-Semantic\\ (MP+U)\end{tabular} & 2.1082                     & 0.0031                     & 0.0016                     & 0.9945                     & 37.5222                     & 37.7290                     \\
				& SPMD                                                               & 2.1629                     & 0.0032                     & 0.0017                     & 0.9940                     & 37.2778                     & 37.5754                     \\
				\multirow{-6}{*}{\begin{tabular}[c]{@{}c@{}}Intermittent\\ Movement\end{tabular}} & \textbf{Our approach} & \textcolor[rgb]{1,0,0}{1.6250} & \textcolor[rgb]{1,0,0}{0.0012} & \textcolor[rgb]{1,0,0}{0.0000} & \textcolor[rgb]{1,0,0}{0.9957} & \textcolor[rgb]{1,0,0}{38.4293} & \textcolor[rgb]{1,0,0}{38.7184}\\
				\hline	
			\end{tabular}

			\begin{tablenotes}
				\footnotesize
				\item[*] Note that \textcolor[rgb]{1,0,0}{red entries} indicate the best in metric.
				
			\end{tablenotes}
			
		\end{threeparttable}
		
	}
\end{table*}

\subsection{Motion Detection Combined with Superpixels}
\label{ssec:correlation}
Superpixel segmentation has attracted the interest of many computer vision applications as it
provides an effective strategy to estimate image features and reduce the complexity of subsequent image processing tasks\cite{wang2017superpixel}. Superpixels have been applied in various fields including object recognition\cite{lu2013superpixel,giordano2015superpixel}, image segmentation\cite{lei2018superpixel} and object tracking\cite{yang2014robust}. 

As most optical flow techniques assumed \cite{baker2011database} that the motion field near motion boundary between foreground and background tend to be over-smoothed and blurred. Motion boundaries are the most important regions and incorrect motions near the area often lead a incorrect result in motion estimation. For effective motion estimation in a scene, we introduce the superpixel segmentation algorithm in the proposed algorithm to further acquire and differentiate the spatial texture information of foreground and background\cite{lim2014generalized,xu2019robust}. Here, SLIC algorithm\cite{achanta2012slic} is utilized on account of its low  complexity and high memory efficiency in computation.

The steps of motion detection are as follows:

\begin{enumerate}
	\item To record the pixels $\left\lbrace p(x_i,y_j) \right\rbrace $ of the foreground detected by CPB;
	\item To estimate the value $V$ of superpixel regions $S$ in these pixels  $\left\lbrace p(x_i,y_j) \right\rbrace $;
	\item Then, to detect the motion and acquire the motion mask $M$, when $\forall\{p(x, y)\} \text{in current frame}$ is denoted as:
	
	\begin{equation}\label{mask}     
     	m(x, y)=\left\{\begin{array}{ll}{1} & {\text { if } p(x, y) \in V} \\ {0} & {\text { otherwise }}\end{array}\right..
	\end{equation}
	
\end{enumerate}

The motion mask $M=\left\lbrace m(x, y)\right\rbrace $. With the help of superpixel segmentation, the proposed approach can further acquire the spatial information of each pixel and distinguish the different motion information between foreground and background. Based on this, the proposed approach can reinforce the original CPB for extracting motion and avoid errors in information extraction from pixels.

\subsection{Final Background Generation}
\label{ssec:generation}
Then, we replace the region of motion mask with the initial CPB background model  for background generation as shown in Fig.~\ref{fig:fig2}.


\section{Experiments}
\label{sec:experimental}

\subsection{Experiment Setup}
\label{ssec:Setup}
In order to fairly evaluate the proposed approach without losing generality, we consider the several challenges in the background initialization algorithm\cite{SBMnet2016}. The following challenges are selected from\textit{ SBMnet} for evaluation:

\begin{itemize}
	
	\item \textbf{Basic:} \textit{PETS2006} represents a mixture of mild challenges typical of the shadows and intermittent movement.
	\item \textbf{Illumination changes:} \textit{Dataset3Camera2} with the illumination changes during day.
	
	\item \textbf{Background motion:} \textit{advertisementBoard} contains an ever-changing advertising board in the scene.
	\item \textbf{Camera jitter:} \textit{boulevard} contains the videos captured by outdoor unstable cameras.
	
	\item \textbf{Intermittent movement:} \textit{sofa} sequence with abandoned objects moving, then stopping for a short while, and then moving again.
	
\end{itemize}

\subsection{Evaluation Measurement}
\label{sec:measurement}
Six metrics which are the common measurements for the background initialization algorithm \cite{SBMnet2016, xu2019robust} are introduced for performance evaluation in this paper. They are explained as follows:

\begin{itemize}
	
	\item \textbf{AGE} (\textit{Average Gray-level Error}): average of the absolute difference between \textit{GT} and \textit{BI}.
	
	\item \textbf{pEPs} (\textit{Percentage of Error Pixels}): number of pixels in \textit{BI} whose value differs from the value of the corresponding pixel in \textit{GT} by more than a threshold $\tau$, which is set as 20 in\cite{SBMnet2016}.
	
	\item \textbf{pCEPs} (\textit{Percentage of Clustered Error Pixels}): percentage of \textit{CEPs} (number of pixels whose 4-connected neighbors are also error pixels) with respect to the total number of pixels in the image.
	
	\item \textbf{PSNR} (\textit{Peak Signal to Noise Ratio}): widely used to measure the quality of  \textit{BI} compared with \textit{GT}, defined as $PSNR=10\cdot{\log}_{10}\left(\dfrac{255^2}{MSE}\right)$.
	
	\item \textbf{MS-SSIM} (\textit{Multi-scale Structural Similarity Index}): estimate of the perceived visual distortion defined in \cite{wang2003multiscale}.
	
		\item \textbf{CQM} (\textit{Color image Quality Measure}): defined in \cite{yalman2013new}. 
			It assumes values in \textit{db} and the higher \textit{CQM} value, the better is the background estimate.
	
\end{itemize}
 
 Where, \textit{GT} means the ground truth of the background image and \textit{BI} means the generated background image computed by the background initialization approaches.

\subsection{Result Evaluation}
\label{sec:Evaluation}
In this section, the proposed approach is compared with five different state-of-the-art techniques selected from SBMnet benchmark, which are LaBGen-OF\cite{laugraud2017memoryless}, MSCL\cite{javed2017background}, FSBE\cite{djerida2019robust}, LaBGen-P-Semantic(MP+U) \cite{laugraud2018labgen} and SPMD\cite{xu2019robust}. Four of them are the leading techniques for background initialization in \textit{SBMnet} benchmark, especially MSCL\cite{javed2017background} which is the top ranked techniques at present. All the results of the five different techniques come from \textit{SBMnet} benchmark.

In experiments, we set each block as $8\times8$ pixels with input frame size of $320{\times}240$ for CPB. All used parameters are listed in Table~\ref{Tab:Tab2}, and a detailed discussion of parameters can be found in\cite{zhou2017visual}. Experimental results of the background initialization are presented in Fig.~\ref{fig:fig4}, and Table~\ref{Tab:Tab3} lists the overall evaluation of these approaches in different challenges. It can be seen from the above results as shown in Fig.~\ref{fig:fig4} and Table~\ref{Tab:Tab2}, that our approach outperforms other techniques in challenges of  \textit{Basic} and \textit{Intermittent Movement}, and for \textit{Background Motion}, our approach has a close performance to LaBGen-P-Semantic(MP+U), which is the best in this challenge. For other two different challenges, our approach also leads the intermediate level compared with other techniques and the performance is acceptable. The comparison shows that our approach is robust and effective for background initialization in different challenges.

The processing time for background initialization is close to 0.15 seconds with frame size of $320{\times}240$ in MATLAB platform (Intel i7 2.40 GHZ and 16G).

\begin{table}[h]
	\renewcommand{\arraystretch}{}
	\caption{{Parameters setting in CPB}}
	\label{Tab:Tab2}
	\centering
	\scalebox{1.0}[1.0]{
		\begin{tabular}{cc}

			\hline
			{Supporting blocks number $K$ } &  {20}\\
			{Threshold of Gaussian model $\eta$} & {2.5}\\
			{Threshold of correlation dependent decision $\lambda$} & {0.5}\\
			\hline

		\end{tabular} %
	}
\end{table}


\section{Conclusions}
\label{sec:conclusion}
In this paper, we propose a new approach for robust background initialization  of a complex scene based on co-occurrence background model (CPB) with superpixel segmentation. It is designed to handle the severe challenges in background initialization, such as illumination changes, background motion, camera jitter and intermittent movement, etc. Video sequences contain the temporal context information which can be learned by CPB model from the training data to resist interference in the scene. Furthermore, superpixel segmentation can help acquire more spatial texture information to facilitate the motion differentiation between foreground and background. The experimental results under different challenges validate the comprehensive performance of the proposed approach. More details including source code are released in: \textit{https://github.com/zwj1archer/CPB-superpixel.git}.


\section*{Acknowledgment}
This work is supported by scientific research starting project of SWPU (No.2019QHZ017).



\bibliographystyle{IEEEtran}
\bibliography{refs}
%



\end{document}